\documentclass[runningheads]{llncs}
\usepackage[T1]{fontenc}
\usepackage{graphicx,verbatim}
\usepackage{hyperref}
\hypersetup{colorlinks=true, citecolor=blue, linkcolor=blue, urlcolor=blue}
\usepackage{color}

\usepackage{amsmath}
\numberwithin{equation}{section}
\usepackage{subcaption}
\usepackage{caption}
\usepackage{soul}
\usepackage{tikz}
\usetikzlibrary{calc} 
\usepackage{float}
\usepackage{orcidlink}
\usepackage{amsmath}
\numberwithin{equation}{section}
\usepackage{float}
\usepackage{longtable}
\usepackage{booktabs}
\usepackage{makecell}

\makeatletter
\newif\if@anonymize

\@anonymizefalse  

\if@anonymize
  \newcommand{\highlight@DoHighlight}{
    \fill [outer sep = -15pt, inner sep = 0pt, color=black]
          ($(begin highlight)+(0,8pt)$) rectangle ($(end highlight)+(0,-3pt)$) ;
  }

  \newcommand{\highlight@BeginHighlight}{
    \coordinate (begin highlight) at (0,0) ;
  }

  \newcommand{\highlight@EndHighlight}{
    \coordinate (end highlight) at (0,0) ;
  }

  \newdimen\highlight@previous
  \newdimen\highlight@current
  \newlength{\item@width}

  \DeclareRobustCommand*\anonymize{%
    \SOUL@setup
    \def\SOUL@preamble{%
      \begin{tikzpicture}[overlay, remember picture]
        \highlight@BeginHighlight
        \highlight@EndHighlight
      \end{tikzpicture}%
    }%
    \def\SOUL@postamble{%
      \begin{tikzpicture}[overlay, remember picture]
        \highlight@EndHighlight
        \highlight@DoHighlight
      \end{tikzpicture}%
    }%
    \def\SOUL@everyhyphen{%
      \discretionary{%
        \SOUL@setkern\SOUL@hyphkern
        \SOUL@sethyphenchar
        \tikz[overlay, remember picture] \highlight@EndHighlight ;%
      }{%
      }{%
        \SOUL@setkern\SOUL@charkern
      }%
    }%
    \def\SOUL@everyexhyphen##1{%
      \SOUL@setkern\SOUL@hyphkern
      \settowidth{\item@width}{##1}%
      \makebox[\item@width]{}%
      \discretionary{%
        \tikz[overlay, remember picture] \highlight@EndHighlight ;%
      }{%
      }{%
        \SOUL@setkern\SOUL@charkern
      }%
    }%
    \def\SOUL@everysyllable{%
      \begin{tikzpicture}[overlay, remember picture]
        \path let \p0 = (begin highlight), \p1 = (0,0) in \pgfextra
          \global\highlight@previous=\y0
          \global\highlight@current =\y1
        \endpgfextra (0,0) ;
        \ifdim\highlight@current < \highlight@previous
          \highlight@DoHighlight
          \highlight@BeginHighlight
        \fi
      \end{tikzpicture}%
      \settowidth{\item@width}{\the\SOUL@syllable}%
      \makebox[\item@width]{}%
      \tikz[overlay, remember picture] \highlight@EndHighlight ;%
    }%
    \SOUL@
  }
\else
  \newcommand{\anonymize}[1]{#1}
\fi
\makeatother

\begin{document}
\title{CLAIRE-DSA: Fluoroscopic Image Classification for Quality Assurance of Computer Vision Pipelines in Acute Ischemic Stroke}
\titlerunning{CLAIRE-DSA}

\author{Cristo J. van den Berg\textsuperscript{*}\inst{1}\orcidlink{0009-0007-6669-4761}\and Frank G. te Nijenhuis\textsuperscript{*}\inst{2}\orcidlink{0009-0003-1321-5836} \and Mirre J. Blaauboer \inst{1} \and Daan T. W. van Erp\inst{1}\orcidlink{0009-0002-0344-0936} \and Carlijn M. Keppels \inst{1} \and Matthijs van der Sluijs\inst{2}\orcidlink{0000-0002-4934-0933}
\and Bob Roozenbeek \inst{1}\orcidlink{0000-0002-8320-8303} \and Wim van Zwam\inst{3}\orcidlink{0000-0003-1631-7056}\and Sandra Cornelissen\inst{2}\orcidlink{0000-0002-0332-2158}\and Danny Ruijters\inst{6}\orcidlink{0000-0002-9931-4047}\and Ruisheng Su\inst{6}\orcidlink{0000-0002-5013-1370}\and Theo van Walsum\inst{2}\orcidlink{0000-0001-8257-7759}} 
\institute{
TU Delft, Mekelweg 5, 2628 CD Delft, The Netherlands\\
\email{\{c.j.vandenberg,m.j.blaauboer,d.t.w.vanerp,c.m.keppels\}@student.tudelft.nl}
\and
Erasmus MC, Doctor Molewaterplein 40, 3015 GD Rotterdam, The Netherlands\\
\email{\{f.tenijenhuis,p.m.vandersluijs,b.roozenbeek,s.cornelissen,t.vanwalsum\}@erasmusmc.nl}
\and
Maastricht UMC+, P. Debyelaan 25, 6229 HX Maastricht, The Netherlands\\
\email{w.van.zwam@mumc.nl}
\and
TU Eindhoven, Groene Loper 5, 5612 AE Eindhoven, The Netherlands\\
\email{\{d.ruijters,r.su\}@tue.nl}\\
\textsuperscript{*}These authors contributed equally to this work.}
\authorrunning{C. J. van den Berg \textit{et al.}}

\maketitle              
\begin{abstract}
Computer vision models can be used to assist during mechanical thrombectomy (MT) for acute ischemic stroke (AIS), but poor image quality often degrades performance. This work presents CLAIRE-DSA, a deep learning–based framework designed to categorize key image properties in minimum intensity projections (MinIPs) acquired during MT for AIS, supporting downstream quality control and workflow optimization. CLAIRE-DSA uses pre-trained ResNet backbone models, fine-tuned to predict nine image properties (e.g., presence of contrast, projection angle, motion artefact severity). Separate classifiers were trained on an annotated dataset containing 1,758 fluoroscopic MinIPs. The model achieved excellent performance on all labels, with ROC-AUC ranging from 0.91 to 0.98, and precision ranging from 0.70 to 1.00. 
The ability of CLAIRE-DSA to identify suitable images was evaluated on a segmentation task by filtering poor quality images and comparing segmentation performance on filtered and unfiltered datasets. Segmentation success rate increased from 42\% to 69\%, $p < 0.001$.
CLAIRE-DSA demonstrates strong potential as an automated tool for accurately classifying image properties in DSA series of acute ischemic stroke patients, supporting image annotation and quality control in clinical and research applications. Source code is available \href{https://gitlab.com/icai-stroke-lab/wp3_neurointerventional_ai/claire-dsa}{here}.



\end{abstract}
\section{Introduction}
Acute ischemic stroke (AIS) is a medical condition characterized by the sudden onset of neurological deficits caused by the occlusion of a cerebral artery, due to which blood flow to the brain is interrupted \cite{Lui2025}. In 2020 alone, 68.15 million people worldwide suffered from AIS, leading to significant mortality and morbidity \cite{Capirossi2023,Lui2025}. 
Endovascular thrombectomy (EVT), is a minimally invasive treatment for AIS. By removing the occlusion from the cerebral artery with a stent-retriever, EVT restores blood flow and significantly improves outcomes \cite{Saver2016}. 

Fluoroscopy, a continuous, real-time X-ray technique, enables dynamic visualization during catheter and guidewire navigation. To visualize vascular anatomy during the procedure, Digital Subtraction Angiography (DSA) images are acquired. DSA is a subtype of fluoroscopy in which contrast medium is injected after acquiring one pre-contrast X-ray image, which serves as a mask. The mask is subtracted from the post-contrast images, leaving only high-resolution vascular images \cite{Shaban2022}. 
Artificial intelligence (AI) has become an expanding field within stroke care, showing promising results in DSA image analysis, such as automated TICI scoring \cite{Su2021}, perforation detection \cite{Su2022} and detection of arterial landmarks and vascular occlusions \cite{Khankari2023}. Given the fact that EVT must be performed immediately, AI could assist in multiple ways in the time-sensitive clinical workflow. 

Clinical implementation of such AI models remains limited due to variable data quality, as DSA is manually acquired and prone to artefacts such as motion \cite{Ruisheng2024}. These artefacts degrade model performance and complicate dataset curation. In recent work by Su \textit{et al.}, for example, 307 out of 987 eligible patients were excluded from further processing, due to either poor image quality or incorrect properties \cite{Su2021}. As a result, dataset preparation becomes labor-intensive and often leads to studies with relatively small sample sizes. Automatic classification of image properties and artefacts helps ensure DSA series are amenable to further analysis.

Oksuz et al.~\cite{OKSUZ2021105909} developed a deep learning framework to detect and correct motion artefacts in brain MRI using synthetic \textit{k}-space corruption and a residual U-net. Their work demonstrated that artefact handling improves both image quality and downstream stroke segmentation. While their method targets MRI, the underlying problem, artefacts degrading automated analysis, is equally relevant in other image modalities. A thorough search of the literature did not yield any articles related to DSA quality assessment, highlighting a potential research gap.

The purpose of this study is to develop a model that performs automatic classification based on image properties in DSA image series of patients suffering from acute ischemic stroke. The main contributions are:
\begin{itemize}
  \item We introduce \emph{Classification AI for Radiological Exams-DSA} (CLAIRE-DSA), a deep learning model designed to automatically classify fluoroscopic images based on nine image properties (e.g.\ \textit{Skull Visibility}, \textit{Projection}, \textit{Motion Artefacts}).
  \item We evaluate CLAIRE-DSA's classification performance and integrate it as a data filter for a downstream segmentation task, evaluating its performance in a realistic application.
  \item We release our entire pipeline, including trained models, as an open-source Python package 
  \href{https://gitlab.com/icai-stroke-lab/wp3_neurointerventional_ai/claire-dsa}{here}.
\end{itemize}

\section{Methods}

\subsection{Model Architecture}
\begin{figure}[t]
    \centering    \includegraphics[width=1\textwidth]{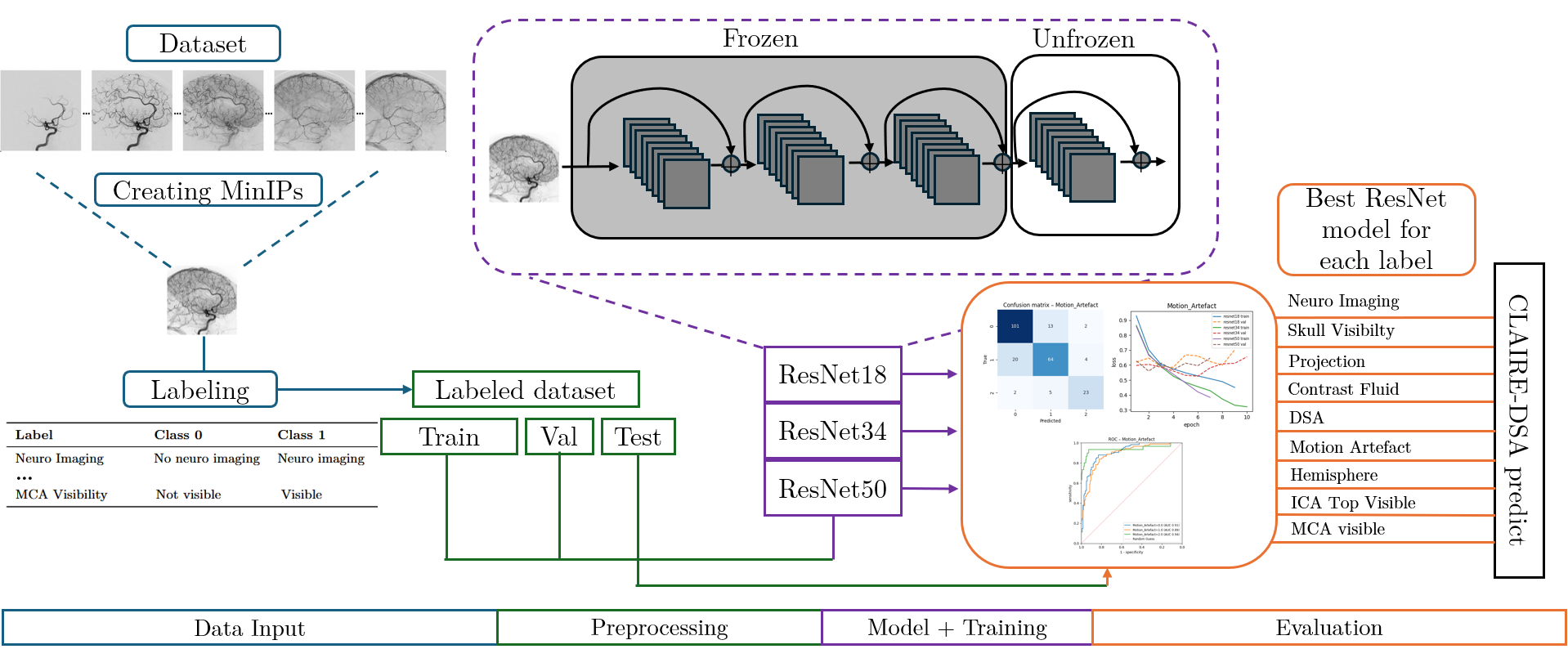}
    \caption{Schematic overview of the model pipeline, illustrating data input and pre-processing, backbone architecture and model training and evaluation. First, MinIPs are created, which are then labeled by four raters and split into train, test and validation sets. For each of nine DSA labels, three pretrained ResNet models are finetuned, and the best performing model is selected. CLAIRE-DSA combines the 9 best models. }
    \label{fig:diagram paper}
\end{figure}

A supervised deep learning approach was used, based on the Residual Neural Network (ResNet) as the backbone. ResNet-based architectures use residual connections, which allow information to skip individual convolutional layers, enabling the creation of deep models \cite{He2015}. For each label, a separate ResNet model was fine-tuned after being pretrained on ImageNet \cite{imagenet}, resulting in nine submodels in total. Each ResNet was partially frozen: only the final three residual blocks were fine-tuned per label. Fig. \ref{fig:diagram paper} shows a schematic overview of the model pipeline.

\begin{table}[t]
\centering
\small
\caption{Overview of annotated labels and their corresponding classes, in parenthesis the number of samples per class. \textit{Left Lat.} and \textit{Right Lat.} refer to a lateral view with the eyes looking left or right. \textit{Hemi}: hemisphere. \textit{Indeterm.}: indeterminable.}
\setlength{\tabcolsep}{3pt}
\makebox[\textwidth][c]{%
\begin{tabular}{llllll}
\toprule
\textbf{Label}     & \textbf{Class 0} & \textbf{Class 1} & \textbf{Class 2} & \textbf{Class 3}\\
\midrule
Neuro Imaging & Not Neuro (166) & Neuro (1592) & - & -  \\
Skull Visibility & Neck (137) & Full (426) & Partial (1029) & - \\
Projection & AP view (939) & Oblique (95) & Left Lat. (93) & Right Lat. (461) \\ 

Contrast Fluid & Absent (272) & Present (1321) & - & -  \\
DSA & Not DSA (285) & DSA (1308) & - & - \\
Motion Artefact & None (792) & Mild (440) & Severe (358) & -  \\
Hemisphere & Indeterm. (792) & Left Hemi (440) & Right Hemi (358) & - \\
ICA Top Visible & Not visible (560) & Visible (1030) & - & -  \\
MCA Visible & Not visible (951) & Visible (638) & - & -  \\
\bottomrule
\end{tabular}}
\label{tab:labels_and_classes}
\end{table}     

\subsection{Labeling \& Model Evaluation}
\label{evaluation procedure}
Four raters (\anonymize{DvE, MB, CK, CvdB}) rated all images. Nine labels were defined to capture key image properties and imaging artefacts that potentially impact the performance and reliability of computer vision models used for processing fluoroscopic images. The labels were defined in consultation with clinical and imaging experts, aiming to capture the most common and relevant properties in fluoroscopic neuro-imaging. All labels and classes are displayed in Tab. \ref{tab:labels_and_classes}.
Fifteen percent of the dataset was reserved as a test set. To determine the best-performing model per label, ResNet-18, ResNet-34, and ResNet-50 backbones were compared, selecting the architecture with the lowest validation loss. For each model, performance was assessed per label using ROC-AUC (binary) or macro ROC-AUC (multiclass), accuracy, precision, recall and $F_1$-score. To gain further insight into the model predictions, we generated model activation maps using Gradient-weighted Class Activation Mapping (Grad-CAM) \cite{Selvaraju_2019}.

\subsection{Label Agreement}
To evaluate labeling consistency between four inexperienced raters and an experienced rater (\anonymize{FtN}), a randomly sampled subset of 126 images was independently labeled by all five raters, based on a minimal sample size formula for $\kappa$ \cite{Donner1992}. Agreement between the experienced rater and the others was assessed with Cohen’s $\kappa$ \cite{McHugh2012}, while overall agreement between raters was evaluated using Fleiss’ $\kappa$ \cite{moons2023}. An agreement $\geq 0.81$ was considered reliable \cite{McHugh2012}.

\subsection{Downstream Task} \label{sec:downstream}
To evaluate the applicability of CLAIRE-DSA in an image processing pipeline, the holdout test set was processed by the Cerebral Artery-Vein Segmentation (CAVE) model, which segments cerebral vessels in DSA images \cite{Ruisheng2024}. An experienced rater (\anonymize{FtN}) classified vessel segmentations as successful or unsuccessful. Successful segmentations were defined as having at most 20\% incorrectly classified pixels upon visual assessment. The same test set was input into CLAIRE-DSA to predict segmentation suitability. Images were deemed unsuitable if any of the following labels were detected: \textit{Neuro Imaging} = Not Neuro, \textit{Skull Visibility} = Neck, \textit{Contrast Fluid} = Not visible, \textit{DSA} = Not DSA image or \textit{Motion Artefact} = Mild or Severe. The primary outcome was the successful segmentation rate in both the unfiltered and CLAIRE-DSA–filtered datasets. A \textit{Z}-test for proportions was used to compare the segmentation success rates between the unfiltered test set and the CLAIRE-DSA filtered test set. A True Positive (TP) is defined as an image that is included by CLAIRE-DSA and correctly segmented. A True Negative (TN) constitutes an image classified as unsuitable with a failed segmentation. False negative (FN) is an image which was excluded by CLAIRE-DSA, but which was still correctly segmented. A False Positive (FP) is an included image that is incorrectly segmented. 
 

\subsection{Implementation}
All code was written in Python. Training was performed on an Nvidia RTX 3060TI GPU. All models were fine-tuned for 10 epochs. The decision to fine-tune for 10 epochs was based on the loss curves. Cross-entropy was used as a loss function in both the binary and multiclass cases. AdamW~\cite{Loshchilov2017} was used as the optimizer, with a weight decay of $1 \times 10^{-4}$. To address class imbalance, dynamic class weights were computed based on the inverse frequency of each class in the training set. To prevent overfitting, a dropout rate of 0.2 was applied.     

\section{Results}

\begin{figure}[t]
    \centering
    \includegraphics[width=0.9\textwidth]{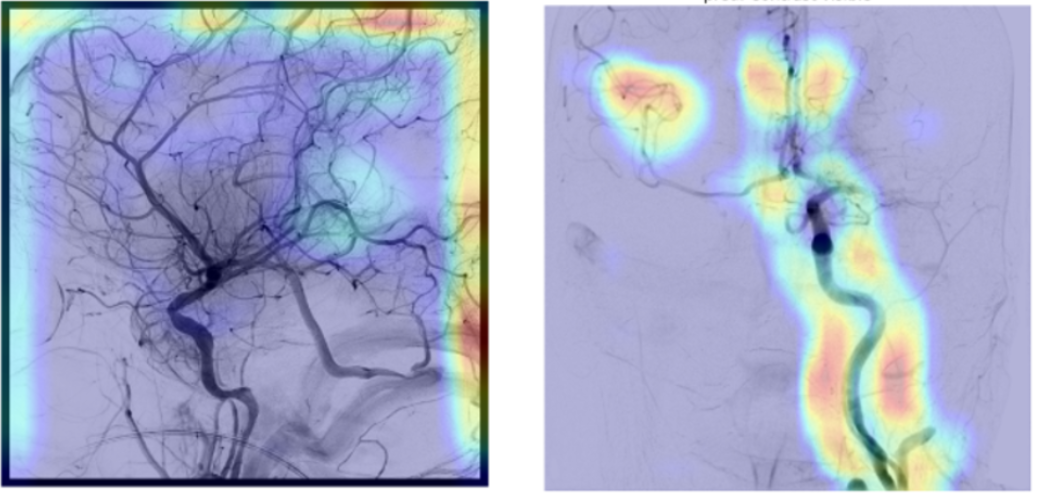}
    \caption{Selected Grad-CAM heatmaps for the \textit{Skull Visibility} (left) and \textit{Contrast Fluid} (right) labels, showing model attention regions based on the gradient output. For \textit{Skull Visibility}, the model clearly looks at the edges of the image to see whether the entire skull is present. For \textit{Contrast Fluid}, the model attends to contrast, as expected.}
    \label{fig:heatmap}
\end{figure}
Data from the \anonymize{MR CLEAN (Multicenter Randomized Clinical trial of Endovascular Treatment for Acute Ischemic Stroke in the Netherlands) Registry} was used. This dataset contains imaging of all patients treated with EVT for acute ischemic stroke in all interventional stroke centres in \anonymize{the Netherlands}. Data collection took place between \anonymize{March 2014 and November 2017}.
We randomly extracted periprocedural fluoroscopic images of 148 randomly sampled patients, yielding 1758 fluoroscopic DICOM sequences. All DICOM series were converted to minimal intensity projections (MinIPs) and normalized to a range of $[0,1]$. The dataset was stratified at the patient level into training, validation, and test sets (70\%/15\%/15\%, respectively), ensuring an adequate balance of labels between the sets. 

\begin{figure}[t]
    \centering
    \begin{subfigure}[b]{0.49\linewidth}
        \includegraphics[width=\linewidth]{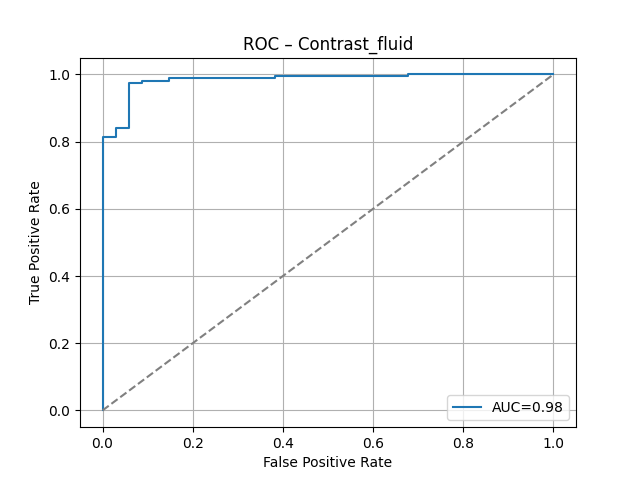}
        \caption{Contrast fluid}
    \end{subfigure}
    \hfill
    \begin{subfigure}[b]{0.49\linewidth}
        \includegraphics[width=\linewidth]{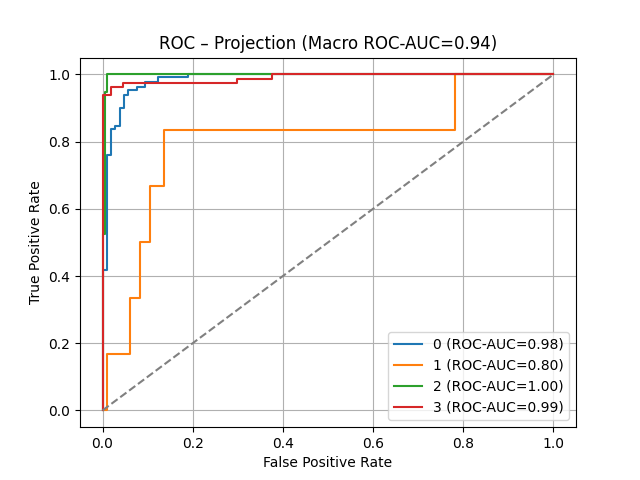}
        \caption{Projection}
    \end{subfigure}
    \caption{Selected ROC curves for the detection of \textit{Contrast fluid} and \textit{Projection}. For contrast, CLAIRE-DSA attains excellent performance (ROC-AUC=0.98), for projection direction, overall performance is still strong (macro-AUC=0.94), with detection of oblique views appearing most challenging with ROC-AUC=0.80.}
    \label{fig:roc}
\end{figure}

For binary labels, ROC-AUC ranged from 0.91 \textit{MCA Visibility} and \textit{Motion Artefact} to 0.98 for \textit{Contrast fluid} and \textit{DSA} (0.98), \textit{ICA Top visible} (0.96), and \textit{MCA visible} (0.91). Multi-class labels showed macro ROC-AUCs ranging from 0.98 for the \textit{Hemisphere} label to 0.91 for \textit{Motion Artefact}. Individual class ROC-AUCs ranged from 0.80 to 1.00. Performance metrics for each label, based on the ResNet variant with the lowest validation loss per label, are shown in Tab.~\ref{tab:avg_metrics_resnet34}. Fig.~\ref{fig:roc} shows the ROC curves and corresponding ROC-AUC values for the \textit{Contrast Fluid} and \textit{Projection} labels, indicating the model performance on a binary and multiclass classification task.

Accuracy of CLAIRE-DSA ranged from 0.80 (\textit{Motion Artefact}) to 0.99 (\textit{DSA}). The highest precision (1.00) was achieved for \textit{DSA}; the lowest precision (0.70) was achieved for \textit{Projection}. Recall ranged from 0.99 (\textit{Contrast Fluid}, \textit{DSA}, and \textit{ICA Top Visible}); to 0.70, again for \textit{Projection}. $F_1$ scores similarly ranged from 0.70 (\textit{Projection}) to 0.99 (\textit{DSA}).

\begin{table}[t]
\centering
\caption{Performance metrics per classification label, after selecting the ResNet backbone with lowest validation loss per label. All models achieve AUC > 0.90, indicating excellent performance.}
\begin{tabular}{llccccc}
\toprule
\textbf{Label} & \textbf{Model} & \textbf{ROC-AUC} & \textbf{Accuracy} & \textbf{Precision} & \textbf{Recall} & \textbf{${F_1}$} \\
\midrule
Neuro Imaging     & ResNet-18  & 0.93 & 0.98 & 0.93 & 0.92 & 0.93 \\
Skull Visibility  & ResNet-34  & 0.96 & 0.87 & 0.83 & 0.86 & 0.84 \\
Projection        & ResNet-34  & 0.94 & 0.93 & 0.70 & 0.70 & 0.70 \\
Contrast Fluid    & ResNet-34  & 0.98 & 0.95 & 0.96 & 0.99 & 0.97 \\
DSA               & ResNet-34  & 0.98 & 0.99 & 1.00 & 0.99 & 0.99 \\
Motion Artefact   & ResNet-34  & 0.91 & 0.80 & 0.80 & 0.79 & 0.79 \\
Hemisphere        & ResNet-34  & 0.98 & 0.89 & 0.89 & 0.87 & 0.88 \\
ICA Top Visible   & ResNet-34  & 0.96 & 0.91 & 0.89 & 0.99 & 0.94 \\
MCA Visible       & ResNet-18  & 0.91 & 0.84 & 0.81 & 0.78 & 0.79 \\
\bottomrule
\end{tabular}
\label{tab:avg_metrics_resnet34}
\end{table}

\begin{table}[t]
\centering
\caption{Confusion matrix with sensitivity and specificity for the clinical evaluation using CAVE. Columns refer to expert assessment of the segmentations, while rows refer to CLAIRE-DSA classification of the images.}
\begin{tabular}{llll}
\toprule
 & \textbf{Correct} & \textbf{Incorrect}\\
\midrule
\textbf{Usable}  & 60 (TP) & 27 (FP)\\
\textbf{Unusable}    & 37 (FN)  & 110 (TN)\\
\midrule
                 & Sensitivity = 0.62 & Specificity = 0.80 & \\
\bottomrule
\end{tabular}
\label{tab:CAVE}
\end{table}

\begin{figure}[t]
  \centering

  \begin{subfigure}[t]{0.2\textwidth}
    \centering
    \includegraphics[width=\linewidth]{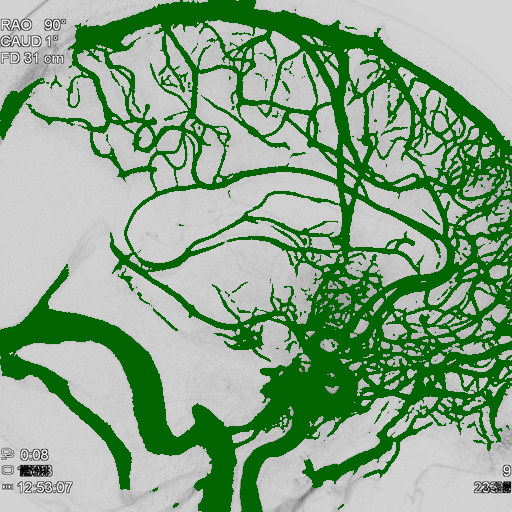}
    \caption*{\scriptsize\textbf{(a) TP}\\CLAIRE (+)\\CAVE (+)}
  \end{subfigure}
  \hspace{1em}
  \begin{subfigure}[t]{0.2\textwidth}
    \centering
    \includegraphics[width=\linewidth]{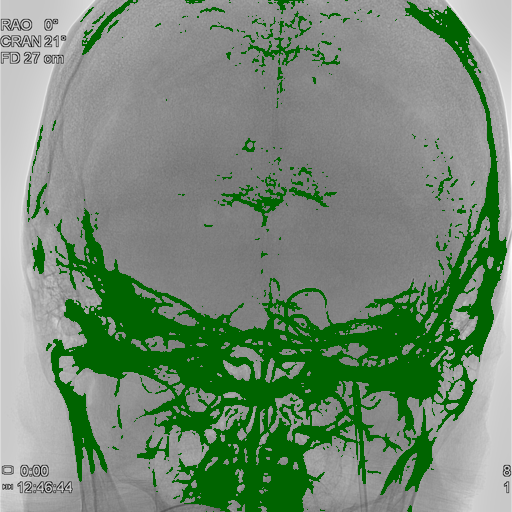}
    \caption*{\scriptsize\textbf{(b) TN}\\CLAIRE (-)\\CAVE (-)}
  \end{subfigure}
  \hspace{1em}
  \begin{subfigure}[t]{0.2\textwidth}
    \centering
    \includegraphics[width=\linewidth]{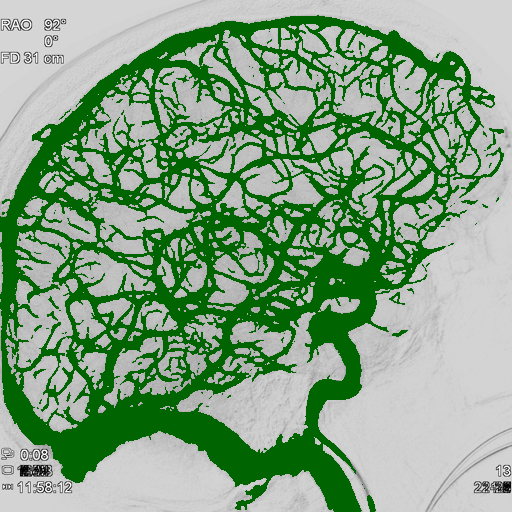}
    \caption*{\scriptsize\textbf{(c) FN}\\CLAIRE (-)\\CAVE (+)}
  \end{subfigure}
  \hspace{1em}
  \begin{subfigure}[t]{0.2\textwidth}
    \centering
    \includegraphics[width=\linewidth]{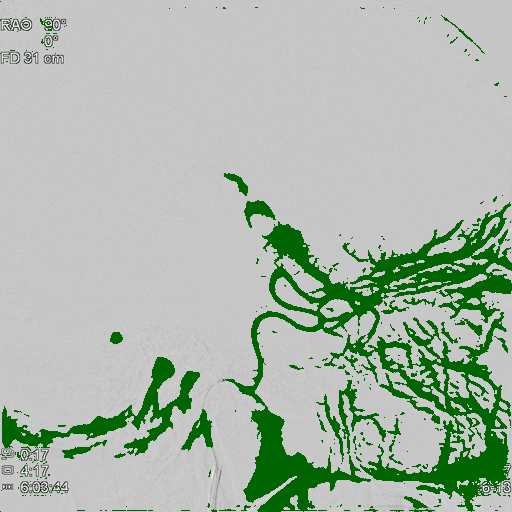}
    \caption*{\scriptsize\textbf{(d) FP}\\CLAIRE (+)\\CAVE (-)}
  \end{subfigure}

  \caption{Example results from the CAVE segmentation and classification. The vessel segmentation is drawn in green on the MinIP background. Note in particular the False Positive \textbf{(d)}, where the DSA image was assessed to be suitable for the downstream task, but vessel segmentation failed, possibly due to lack of contrast medium in the image.}
  \label{fig:example_cave}
\end{figure}
Of the 1,872 classifications made on the test set, 197 were misclassified (11\%), 28 of which were attributable to annotation errors identified in retrospect. Common image characteristics associated with misclassifications included fluoroscopy images labeled as DSA ($n=20$), low image contrast ($n=9$), extreme motion artefacts ($n=11$) and presence of a catheter ($n=6$), affecting multiple labels such as \textit{Skull Visibility}, \textit{Contrast Fluid}, \textit{Hemisphere}, and \textit{Motion Artefact}.

For the inter-observer variability, only the labels \textit{Skull Visibility}, \textit{Motion Artefact}, and \textit{MCA Visible} did not pass the agreement threshold with a Fleiss $\kappa$ score of 0.713, 0.594, and 0.637, respectively. In the label quality analysis, every label reached a Cohen's $\kappa$ $\ge$ 0.81 except \textit{Motion Artefact} and \textit{MCA Visible}, with a $\kappa$ of 0.779 and 0.786, respectively.

For the clinical evaluation with CAVE, the same external testset ($n=234$) used for performance evaluation was labeled by CLAIRE-DSA, segmented by CAVE, and classified by an experienced annotator (\anonymize{FtN}). For the definitions of TP, FP, TN and FN, see Section \ref{sec:downstream}. The successful segmentation rate for the entire test set was 42\%. The filtered test set ($n=87$) permitted an improved success rate of 69\%. A \textit{Z}-test for proportions yielded $p < 0.001$, indicating a significant increase in success rate after applying CLAIRE-DSA.
For the task of detecting suitable images for segmentation, CLAIRE-DSA attained a sensitivity of 62\% and a specificity of 80\%, see Tab. \ref{tab:CAVE}. Fig. \ref{fig:example_cave} shows example results after CAVE segmentation and expert classification.



\section{Discussion}
This study addresses the lack of automated quality assessment tools for fluoroscopy in AIS by introducing CLAIRE-DSA, a multi-label deep learning-based classification model for nine key image properties. CLAIRE-DSA achieved excellent ROC-AUC scores (0.91–0.98) and improved downstream task success rate from 42\% to 69\% ($p<0.001$) by filtering out unsuitable images, demonstrating its practical utility in fluoroscopy preprocessing pipelines. Sensitivity in the downstream task was relatively low (62\%), indicating that some images which would have been correctly segmented were rejected by CLAIRE-DSA. As the labeling performance of CLAIRE-DSA is excellent, this is most likely due to a suboptimal selection of filtering properties. This highlights the importance of adequate label selection when using CLAIRE-DSA to filter for a downstream task. Prior work has demonstrated that a substantial portion of fluoroscopic data must be excluded due to poor image quality before it can be used as input for computer vision models, see for instance Su \textit{et al.} \cite{Su2021}, highlighting a use-case for CLAIRE-DSA either by providing feedback to the clinician during image acquisition or by filtering images after acquisition. 
Grad-CAM based post-hoc visualizations confirmed that model decisions were based on plausible image regions, enhancing interpretability and transparency.
Relatively high inter-rater variability ($\kappa<0.81$) for certain labels, like \textit{Motion Artefact}, \textit{MCA Visibility}, and \textit{Skull Visibility}, likely hindered consistent pattern learning, reflected in lower ROC-AUC scores for these labels. Only one rater conducted both label validation and CAVE evaluation, introducing potential bias. 
From a clinical perspective, CLAIRE-DSA holds potential as a quality control tool that can filter suboptimal images before they are used in automated image analysis pipelines. This could reduce the risk of error in AI-based image interpretation and streamline DSA-based computer vision pipelines. CLAIRE-DSA represents a step toward practical integration of automated image quality control in interventional radiology and stroke care. Future research should explore further integration of CLAIRE-DSA into DSA based computer vision models. 

\section{Conclusion}
CLAIRE-DSA shows promising results when classifying image properties in DSA image series of patients with acute ischemic stroke. The model was able to accurately classify a range of relevant visual features across multiple relevant labels. CLAIRE-DSA also significantly improved the success rate of a DSA segmentation model. The results suggest that CLAIRE-DSA may serve as a tool for automated image annotation or quality control in clinical and research settings.

\begin{credits}

\subsubsection{\ackname} We thank the \hyperlink{https://www.linkedin.com/company/icai-stroke-lab/}{ICAI Stroke Lab} collaborators for their contributions. This publication is part of the project ROBUST: Trustworthy AI-based Systems for Sustainable Growth with project number KICH3.LTP.20.006, which is (partly) financed by the Dutch Research Council (NWO), Philips Medical Systems Nederland B.V., and the Dutch Ministry of Economic Affairs and Climate Policy (EZK) under the program LTP KIC 2020-2023.

\subsubsection{\discintname} The authors have no competing interests to declare that are relevant to the content of this article.
\end{credits}
%
%
%
\bibliographystyle{splncs04}
\bibliography{bibliography.bib}

\end{document}